\documentclass[sigconf]{acmart}

\usepackage{booktabs}
\usepackage{tabularx}
\usepackage{array}
\usepackage{multirow}
\usepackage{graphicx}
\usepackage{xcolor}
\usepackage{colortbl}
\usepackage{algorithm}
\usepackage{algpseudocode}
\usepackage{placeins}

\newcommand{\client}{\mathrm{C}}
\newcommand{\server}{\mathrm{S}}
\newcommand{\target}{\mathrm{F}}
\newcommand{\retain}{\mathrm{R}}

\newcommand{\DF}{D_{\mathrm{F}}}
\newcommand{\DR}{D_{\mathrm{R}}}

\AtBeginDocument{%
  }

\theoremstyle{plain}

\theoremstyle{definition}

\theoremstyle{remark}

\renewcommand\footnotetextcopyrightpermission[1]{}

\begin{document}

% ============================================================
% Title
% ============================================================
% \title{Partition-Constrained Representation Realignment for Split Federated Unlearning}
\title{SplitReAlign: Partition-Aware Representation Realignment for Split Federated Unlearning}

% ============================================================
% Authors (commented out for double-blind review)
% ============================================================

% \author{Jia Chen}
% \authornote{Jia Chen and Mingdi Huang contributed equally to this work.}
% \affiliation{%
%   \department{School of Computer Science and Artificial Intelligence}
%   \institution{Wuhan Textile University}
%   \city{Wuhan 430200}
%   \country{China}}
% \additionalaffiliation{%
%   \institution{Engineering Research Center of Hubei Province for Clothing Information}
%   \city{Wuhan 430200}
%   \country{China}}
% \email{j_chen@wtu.edu.cn}

% \author{Yecheng Lv}
% \authornotemark[1]
% \affiliation{%
%   \department{School of Computer Science and Artificial Intelligence}
%   \institution{Wuhan Textile University}
%   \city{Wuhan 430200}
%   \country{China}}
% \email{2515423002@wtu.edu.cn}

% \author{Xinrong Hu}
% \affiliation{%
%   \department{School of Computer Science and Artificial Intelligence}
%   \institution{Wuhan Textile University}
%   \city{Wuhan 430200}
%   \country{China}}
% \email{hxr@wtu.edu.cn}

% \author{Jun Bai}
% \authornote{Corresponding author.}
% \orcid{0000-0003-3349-5161}
% \affiliation{%
%   \department{School of Computer Science}
%   \institution{McGill University}
%   \city{Montreal, QC H3A 0E9}
%   \country{Canada}}
% \email{jun.bai@mcgill.ca}

\author{%
\textbf{
    Jia Chen\textsuperscript{1,2,*},
    Yecheng Lv\textsuperscript{1,*},
    Xinrong Hu\textsuperscript{1},
    % Mingdi Huang\textsuperscript{1},
    Jun Bai\textsuperscript{3,$\dagger$}
    }
}

\affiliation{%
  \institution{%
  \textsuperscript{1} School of Computer Science and Artificial Intelligence,
  Wuhan Textile University, Wuhan, China\\
  \textsuperscript{2} Engineering Research Center of Hubei Province
  for Clothing Information, Wuhan Textile University,
  Wuhan, China\\
    \textsuperscript{3} Australian Institute for Machine Learning (AIML),
    Adelaide University, Adelaide, Australia \\
  }
  \country{\unskip}
}

% ============================================================
% Abstract
% ============================================================
\begin{abstract}

Split federated learning (SFL) distributes neural networks across clients and a server, coupling class-specific knowledge across their representations. 
% Split federated learning (SFL) distributes neural networks across clients and a server in Web-based collaborative learning, coupling class-specific knowledge across their representations.
This coupling complicates split federated unlearning (SFU), where class-wise forgetting must remove target knowledge while preserving retained utility under heterogeneous representation spaces and asymmetric update constraints. We propose \textbf{SplitReAlign}, a partition-aware representation realignment framework for class-wise SFU. It constructs endpoint-specific discriminative subspaces coordinated through a shared semantic anchor, selectively realigning forgotten representations while penalizing changes outside the active subspaces. To accommodate asymmetric update constraints, SplitReAlign leverages gradients through frozen servers for Client-only updates and fixed client-feature statistics for Server-only guidance, while Joint updates optimize both partitions.
Experiments on CIFAR-10 and CIFAR-100 under heterogeneous client distributions demonstrate that SplitReAlign achieves the lowest average gap to retraining among all compared methods across six settings. Notably, Server-only achieves the closest agreement with retraining on both datasets, highlighting the effectiveness of partition-aware guidance under restricted updates.
% Experiments on CIFAR-10/100 with heterogeneous clients show that SplitReAlign achieves the lowest average gap to retraining across six settings, with Server-only exhibiting the closest agreement on both datasets.

\end{abstract}

% ============================================================
% CCS Concepts
% ============================================================
% Generate and insert the official CCS XML from https://dl.acm.org/ccs
% \ccsdesc[500]{Computing methodologies~Distributed artificial intelligence}
% \ccsdesc[300]{Security and privacy~Privacy-preserving systems}

% ============================================================
% Keywords
% ============================================================

\keywords{Split federated learning,
Split federated unlearning,
Class-wise unlearning,
Representation realignment,
Partition-aware unlearning
}

% ============================================================
% ACM title block
% ============================================================

\maketitle

\pagestyle{plain}
\thispagestyle{plain}
\begingroup
\renewcommand{\thefootnote}{\fnsymbol{footnote}}
\setcounter{footnote}{1}
\footnotetext{Jia Chen and Yecheng Lv contributed equally to this work.}

\setcounter{footnote}{2}
\footnotetext{Corresponding author: Jun Bai
(\texttt{junbai.research@gmail.com}).}
\endgroup

\section{Introduction}
\label{sec:introduction}

Split federated learning (SFL) enables collaborative training by partitioning models across clients and servers~\cite{thapa2022splitfed}.
% Split federated learning (SFL) enables distributed Web applications to collaboratively train partitioned neural networks across clients and servers~\cite{thapa2022splitfed}.
However, removing learned information becomes challenging when knowledge is distributed across coupled model partitions. Split federated unlearning (SFU)~\cite{chen2026sure} further considers restricted update modes, where only client-side, server-side, or both partitions can be modified, complicating the balance between effective forgetting and retained utility.

Existing machine unlearning~\cite{fan2024salun,li2026cross} and federated unlearning methods~\cite{zhao2024mode,pan2025fedosd} primarily rely on parameter-level or gradient-based interventions. SURE~\cite{chen2026sure}, the first dedicated SFU approach, addresses partition-constrained unlearning through importance-guided perturbation and retained-data recovery.
Nevertheless, such parameter-level interventions do not coordinate how forgotten representations evolve across the split. This limitation is particularly relevant as reduced target predictions do not necessarily imply the removal of discriminative information from internal representations~\cite{gao2026illusion}.

Beyond these limitations, SFU presents a fundamental challenge: representation changes are asymmetric across model partitions. Client-side updates affect both representation endpoints, whereas server-side updates leave cut-layer features unchanged. Moreover, the two endpoints occupy distinct representation spaces, making consistent semantic realignment difficult, particularly when one partition is frozen. This raises a central question: \emph{How can forgotten representations be coherently realigned under partition-specific update constraints while preserving retained-data utility?}

We propose \textbf{SplitReAlign}, a partition-aware representation realignment framework for class-wise SFU. It constructs endpoint-specific subspaces coordinated by a shared semantic anchor, selectively realigning forgotten representations while limiting changes to retained features. To accommodate asymmetric update constraints, SplitReAlign combines adaptive loss calibration with cross-partition guidance, leveraging gradients from frozen servers and representation statistics from frozen clients. Experiments on CIFAR-10 and CIFAR-100 demonstrate the lowest retraining gap across all six evaluated settings, consistently outperforming SURE and adapted unlearning baselines. Notably, Server-only achieves the closest agreement with retraining, highlighting the value of partition-aware guidance over increased update flexibility.

% Single-column framework overview.
\begin{figure}[t]
    \centering
    \includegraphics[
        width=\columnwidth,
        height=0.45\textheight,
        keepaspectratio
    ]{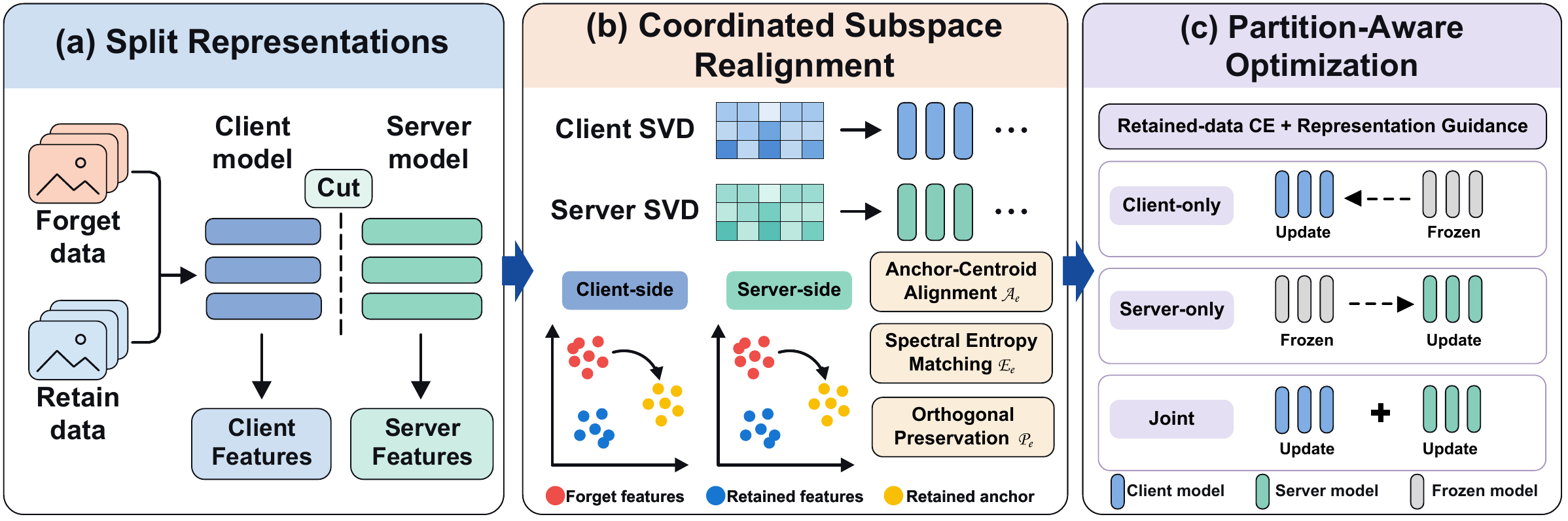}
  
\caption{Overview of SplitReAlign. Endpoint-specific subspaces are coordinated through a shared semantic anchor to realign forgotten representations while preserving shared features. Partition-aware optimization exploits frozen-server gradients in Client-only, fixed-client feature guidance in Server-only, and coordinated updates in Joint.}

    \label{fig:splitrealign_overview}
    \vspace{-0.6em}
\end{figure}

% ============================================================
% 4 Method
% This section presents the proposed SplitReAlign framework,
% including partition-aware representation realignment,
% loss calibration, and mode-constrained unlearning strategies
% for Client-only, Server-only, and Joint updates.
% ============================================================

% \section{Method}
% \label{sec:method}
% SplitReAlign performs class-wise unlearning by reshaping representations under explicit partition-update constraints. Starting from a trained split model $f_\theta=f_{\theta_{\server}}\circ f_{\theta_{\client}}$, it uses the forget set $\DF$, containing target-class training examples across clients, and the retained set $\DR$. A frozen copy of the original model, denoted by $\theta^0$, provides reference representations. Our procedure combines endpoint-specific subspaces, a shared anchor label, and loss-scale calibration, while restricting parameter updates to the partition or partitions permitted by each unlearning mode. The overall framework is illustrated in
% Fig.~\ref{fig:splitrealign_overview}.

\section{SplitReAlign}
\label{sec:method}

% In SFL, parameter updates have asymmetric effects on intermediate representations: updating the client partition can alter both cut-layer and server-side features, whereas updating the server partition leaves cut-layer features unchanged. This asymmetry motivates a partition-aware approach that coordinates representation realignment according to the available update paths. We propose SplitReAlign for class-wise unlearning under client-only, server-only, and joint updates. Given a trained split model $f_\theta=f_{\theta_{\server}}\circ f_{\theta_{\client}}$, we denote the target-class forget set by $\DF$ and the retained set by $\DR$, with a frozen copy $\theta^0$ serving as the reference model. SplitReAlign constructs endpoint-specific discriminative subspaces coordinated by a shared semantic anchor, balances their realignment objectives, and adapts representation guidance to the permitted parameter updates. Fig.~\ref{fig:splitrealign_overview} illustrates the overall framework.

In SFL, client-side updates can alter both cut-layer and server-side representations, whereas server-side updates leave cut-layer features unchanged. This asymmetry motivates \textbf{SplitReAlign}, a partition-aware framework for class-wise unlearning under Client-only, Server-only, and Joint updates. Given a trained split model $f_\theta=f_{\theta_{\server}}\circ f_{\theta_{\client}}$, let $\DF$ and $\DR$ denote the forget and retain sets, respectively, and $\theta^0$ the frozen reference model. As illustrated in Fig.~\ref{fig:splitrealign_overview}, SplitReAlign coordinates endpoint-specific subspaces through a shared semantic anchor, calibrates their realignment objectives, and adapts representation guidance to the permitted updates.

\subsection{Endpoint-Specific Subspaces and Semantic Coordination}
\label{sec:subspace_anchor}

The cut layer and server-side extractor encode class information in distinct feature spaces, making a shared projection unsuitable. Conversely, independent alignment targets may yield inconsistent semantic guidance. SplitReAlign addresses both challenges through endpoint-specific discriminative subspaces coordinated by a shared semantic anchor.

Let $z_{\client}(x)$ denote the spatially averaged cut-layer representation and $z_{\server}(x)$ the pooled server representation before classification. For each endpoint $e\in\{\client,\server\}$, we extract a feature matrix $F_{e,t}$ from a fixed target-class sample at round $t$. Using the retained-feature mean $\bar z^0_{e,\retain}$ from the frozen original model, we perform
\begin{equation}
M_{e,t}
=F_{e,t}-\mathbf{1}(\bar z^0_{e,\retain})^\top
=U_{e,t}\Sigma_{e,t}V_{e,t}^\top.
\end{equation}
The projection basis $P_{e,t}$ comprises leading right singular vectors capturing at least $95\%$ of residual energy, isolating dominant target-to-retain deviations at each endpoint without shared feature coordinates. Bases are refreshed between rounds and frozen when the leading singular value falls below half its initial value. All bases and reference statistics are detached from gradient computation.

To coordinate realignment across endpoints, we select a common retained-class label while maintaining endpoint-specific reference statistics. For each retained class $k$, its projected reference centroid is $\mu^0_{e,k,t}=\mu(Z^0_{e,k}P_{e,t})$, where $Z^0_{e,k}$ contains original reference features and $\mu(\cdot)$ denotes the row mean. At each round, we sample five retained candidate classes $\mathcal{K}_t$ and select
\begin{equation}
k_t=\arg\min_{k\in\mathcal{K}_t}
\left\|\mu(F_{o,t}P_{o,t})-\mu^0_{o,k,t}\right\|_2^2,
\end{equation}
where $o=\client$ for Client-only updates and $o=\server$ for Server-only and Joint updates. Both endpoints use the same anchor label $k_t$, but retain their own projected centroids and dispersion statistics. This maintains semantic consistency across the split while allowing each endpoint to realign representations within its own geometry.

\subsection{Subspace-Guided Representation Realignment}
\label{sec:representation_objective}

Given the endpoint-specific subspaces and shared semantic anchor, SplitReAlign selectively redirects target-class representations within the active subspaces while preserving shared structure in their orthogonal complements. This is achieved through three complementary objectives: centroid alignment, spectral dispersion matching, and orthogonal feature preservation.

At round $t$, let $Z_e$ and $Z_e^0$ denote the current and frozen-reference features of the same augmented minibatch at endpoint $e$. We define $X_e=Z_{e,\target}P_{e,t}$ as the projected forget features and $Y_e=Z^0_{e,k_t}P_{e,t}$ as the projected reference features of the selected anchor class. The realignment objectives are
\begin{align}
\mathcal{A}_e
&=\|\mu(X_e)-\mu(Y_e)\|_2^2,
\label{eq:anchor_loss}\\
\mathcal{E}_e
&=[h(X_e)-h(Y_e)]^2,
\label{eq:entropy_loss}
\end{align}
where $\mathcal{A}_e$ directs the target-class centroid toward the retained anchor, while $\mathcal{E}_e$ regulates the spectral concentration of the projected features. Specifically, $h(\cdot)$ measures the linear entropy of the normalized feature covariance:
\begin{equation}
h(X)=1-\left\|
\frac{X_c^\top X_c}
{\operatorname{tr}(X_c^\top X_c)+\epsilon}
\right\|_F^2,
\quad
X_c=X-\mathbf{1}\mu(X),
\end{equation}
where $\epsilon=10^{-12}$ ensures numerical stability. Entropy matching encourages target representations to exhibit spectral concentration comparable to the selected anchor.

To limit changes to shared representations, we penalize deviations from the original model outside the active subspace:
\begin{equation}
\mathcal{P}_e=
\frac{
\|(Z_e-Z_e^0)(I-P_{e,t}P_{e,t}^{\top})\|_F^2
}{B d_e},
\label{eq:preserve_loss}
\end{equation}
where $B$ is the minibatch size and $d_e$ is the endpoint feature dimension. This term penalizes changes in the orthogonal complement for both forgotten and retained examples, thereby complementing the selective realignment objectives.

The resulting endpoint and overall objectives are
\begin{align}
\mathcal{L}_e
&=\lambda_a\mathcal{A}_e
+\lambda_e\mathcal{E}_e
+\mathcal{P}_e,\\
\mathcal{L}
&=\mathcal{L}_{\mathrm{CE}}^{\retain}
+w_{\client,t}\mathcal{L}_{\client}
+w_{\server,t}\mathcal{L}_{\server},
\label{eq:splitrealign_objective}
\end{align}
where $\mathcal{L}_{\mathrm{CE}}^{\retain}$ preserves classification utility using retained examples and their original labels, while $w_{\client,t}$ and $w_{\server,t}$ balance endpoint objectives. The anchor serves as a representation target rather than a replacement class label. Empty forget batches omit alignment terms, and singleton forget batches omit entropy matching. All reference features and subspace bases remain detached from optimization.

% \subsection{Adaptive Endpoint-Loss Calibration}
% \label{sec:loss_calibration}
% Client-side and server-side representation losses
% may differ substantially in scale. Motivated by the
% inverse-loss-scale balancing principle of
% Hu et al.~\cite{hu2019adaptive},
% we introduce an exponential moving average (EMA)-based calibration mechanism
% that separates endpoint preferences from
% loss-scale correction. Client-only and Joint first use unscaled $0.5{:}0.5$ weights for three warm-up rounds. Let $\ell_{e,t}$ be the round-level unscaled endpoint loss, $m_e^0$ its warm-up average, and $s=(m_{\client}^0+m_{\server}^0)/2$. Initializing $m_{e,3}=m_e^0$, subsequent rounds use
% \begin{align}
% w_{e,t}&=a_e\frac{s}{\max\{m_{e,t-1},\,0.05m_e^0,\,10^{-8}\}},\quad t>3,\\
% m_{e,t}&=0.9m_{e,t-1}+0.1\ell_{e,t}.
% \end{align}
% The fixed reference scale $s$ preserves the warm-up magnitude, while the exponential moving average tracks later changes. The denominator floor limits amplification when an endpoint loss becomes small. If the reference scale is numerically negligible, scale correction is disabled.
% We use the balanced preference
% $a_{\client}=a_{\server}=0.5$ after warm-up.
% The calibrated weights are detached and need not
% sum to one; retained-data cross-entropy is unchanged.

\subsection{Cross-Endpoint Loss Calibration}
\label{sec:loss_calibration}

Although the two endpoints share a semantic anchor, their representation losses may differ in scale, causing one endpoint to dominate the realignment objective. To balance their contributions without altering the objectives, we adopt an EMA-based calibration strategy inspired by inverse-loss-scale weighting~\cite{hu2019adaptive}.

For Client-only and Joint updates, we first use equal endpoint weights for three warm-up rounds. Let $\ell_{e,t}$ denote the unscaled endpoint loss at round $t$, $m_e^0$ its warm-up average, and $s=(m_{\client}^0+m_{\server}^0)/2$. With $m_{e,3}=m_e^0$, subsequent weights are computed as
\begin{align}
w_{e,t}
&=a_e\frac{s}{\max\{m_{e,t-1},\,0.05m_e^0,\,10^{-8}\}},\quad t>3,\\
m_{e,t}
&=0.9m_{e,t-1}+0.1\ell_{e,t},
\end{align}
where $a_{\client}=a_{\server}=0.5$ ensures balanced endpoint preferences, while EMA adjusts for changing loss scales. The denominator floor prevents excessive amplification, and calibration is disabled when $s$ is negligible. Weights are detached and need not sum to one; retained-data classification remains unchanged.

\subsection{Partition-Aware Optimization}
\label{sec:constrained_optimization}

The key distinction of partition-constrained unlearning lies in the asymmetric dependency between model parameters and their representations. Client-side updates can influence both endpoints, whereas server-side updates cannot modify cut-layer features. SplitReAlign exploits this dependency to adapt representation guidance to the trainable partition.

Let $z$ denote the cut-layer activation and define $\mathcal{J}_{\server}=\mathcal{L}_{\mathrm{CE}}^{\retain}+w_{\server,t}\mathcal{L}_{\server}$. The parameter gradients decompose as
\begin{equation}
\begin{aligned}
g_{\client}
&=w_{\client,t}\nabla_{\theta_{\client}}\mathcal{L}_{\client}
+\left(\frac{\partial z}{\partial\theta_{\client}}\right)^\top
\nabla_z\mathcal{J}_{\server},\\
g_{\server}
&=\nabla_{\theta_{\server}}\mathcal{J}_{\server}.
\end{aligned}
\end{equation}
This decomposition reveals that server-side representation objectives can guide client updates through the cut layer without modifying server parameters.

In \emph{Client-only}, only $g_{\client}$ is applied, with the frozen server propagating $\nabla_z\mathcal{J}_{\server}$ to guide client updates through both endpoint objectives. In \emph{Joint}, both gradients are applied for coordinated optimization. In \emph{Server-only}, client parameters are frozen and cut-layer features detached, disconnecting the client-side objective from trainable parameters. Thus, $(w_{\client,t},w_{\server,t})=(0,1)$ is used without calibration.

\paragraph{Client-guided server realignment.}
Although client features are fixed in Server-only mode, they can still inform which forget representations require stronger alignment. We therefore use client-side feature deviations to reweight the server-side forget centroid. For each forget sample $i$, define
\begin{equation}
d_i=\|z_{\client}(x_i)P_{\client,t}
-\mu^0_{\client,k_t,t}\|_2^2,
\qquad r_i=d_i-\bar d,
\end{equation}
where $\bar d$ is the minibatch mean of $d_i$. The detached weights and weighted server centroid are
\begin{equation}
v_i=1+\frac{0.25r_i}
{\max\{\max_j|r_j|,10^{-12}\}},
\quad
\widetilde{\mu}(X_{\server})
=\frac{\sum_i v_iX_{\server,i}}{\sum_i v_i}.
\end{equation}
Replacing the forget centroid in Eq.~\eqref{eq:anchor_loss} with $\widetilde{\mu}(X_{\server})$ assigns greater influence to samples farther from the client-side anchor. The weights remain within $[0.75,1.25]$, while the server-selected anchor, entropy matching, and preservation objectives remain unchanged.

\paragraph{Federated execution.}
The split server maintains a server-side replica per client. Each round refreshes subspaces, performs local split optimization, and separately aggregates trainable partitions using retained-sample weights. Clients without retained data receive zero aggregation weight, while frozen partitions retain their parameters and buffers throughout unlearning.

% ============================================================
% 5 Experimental Evaluation
% Insert this section after \maketitle in the main ACM manuscript.
% Required packages are already included in the main preamble:
% booktabs, graphicx, xcolor, colortbl, array.
% ============================================================

\section{Experimental Evaluation}
\label{sec:experiments}

We evaluate SplitReAlign for class-wise unlearning under three partition-specific update constraints, comparing forgetting effectiveness and retained-class utility against representative baselines under heterogeneous client data distributions.

\subsection{Experimental Setup}
\label{sec:exp_setup}

% \paragraph{Datasets and configurations.}
% We evaluate class-wise unlearning on CIFAR-10 and
% CIFAR-100 using an 18-layer residual network
% (ResNet-18)~\cite{he2016deep} with Group
% Normalization (GN)~\cite{wu2018group}. Specifically, the target class is airplane (class 0) for CIFAR-10 and apple (class 0) for CIFAR-100. The datasets are distributed across ten clients using class-wise Dirichlet ($\alpha=0.1$) partitioning.

\paragraph{Datasets and settings.}
We evaluate CIFAR-10 and CIFAR-100 using ResNet-18. Each dataset is distributed across ten clients using class-wise Dirichlet partitioning ($\alpha=0.1$)~\cite{chen2026sure}. We designate class 0 as the unlearning target.

\providecommand{\SplitReAlignMeanSD}[2]{#1\textsubscript{\tiny$\pm$#2}}

\begin{table*}[t]
\centering
\caption{Class-wise unlearning on CIFAR-10 and CIFAR-100 under
heterogeneous client distributions (Dirichlet $\alpha=0.1$).
Results are mean $\pm$ standard deviation over three seeds.}
\label{tab:class_wise_main}
\scriptsize
\setlength{\tabcolsep}{2.15pt}
\renewcommand{\arraystretch}{0.97}
\resizebox{\textwidth}{!}{%
\begin{tabular}{@{}cl*{15}{c}@{}}
\toprule
& & \multicolumn{5}{c}{Client-only}
  & \multicolumn{5}{c}{Server-only}
  & \multicolumn{5}{c}{Joint} \\
\cmidrule(lr){3-7}
\cmidrule(lr){8-12}
\cmidrule(lr){13-17}
& Method
& Retain & Forget & Test & AG$\downarrow$ & MIA
& Retain & Forget & Test & AG$\downarrow$ & MIA
& Retain & Forget & Test & AG$\downarrow$ & MIA \\
\midrule[1pt]

% ===================== CIFAR-10 =====================

\multirow{10}{*}{\rotatebox[origin=c]{90}{\shortstack[c]{CIFAR-10\\Dirichlet ($\alpha=0.1$)}}}
& Retrain
& \SplitReAlignMeanSD{80.22}{3.62}
& \SplitReAlignMeanSD{0.00}{0.00}
& \SplitReAlignMeanSD{77.44}{2.24}
& 0.00
& \SplitReAlignMeanSD{0.493}{0.009}
& \SplitReAlignMeanSD{80.22}{3.62}
& \SplitReAlignMeanSD{0.00}{0.00}
& \SplitReAlignMeanSD{77.44}{2.24}
& 0.00
& \SplitReAlignMeanSD{0.493}{0.009}
& \SplitReAlignMeanSD{80.22}{3.62}
& \SplitReAlignMeanSD{0.00}{0.00}
& \SplitReAlignMeanSD{77.44}{2.24}
& 0.00
& \SplitReAlignMeanSD{0.493}{0.009} \\

\cmidrule(l){2-17}

& Fine-tune
& \SplitReAlignMeanSD{81.96}{2.75}
& \SplitReAlignMeanSD{44.47}{3.19}
& \SplitReAlignMeanSD{78.78}{2.18}
& 15.85
& \SplitReAlignMeanSD{0.522}{0.025}
& \SplitReAlignMeanSD{82.33}{2.67}
& \SplitReAlignMeanSD{9.73}{11.68}
& \SplitReAlignMeanSD{79.04}{2.10}
& 4.48
& \SplitReAlignMeanSD{0.537}{0.020}
& \SplitReAlignMeanSD{82.77}{2.71}
& \SplitReAlignMeanSD{10.83}{11.92}
& \SplitReAlignMeanSD{79.36}{2.01}
& 5.10
& \SplitReAlignMeanSD{0.534}{0.036} \\

& SalUn
& \SplitReAlignMeanSD{55.69}{10.98}
& \SplitReAlignMeanSD{0.97}{0.91}
& \SplitReAlignMeanSD{54.96}{11.41}
& 15.99
& \SplitReAlignMeanSD{0.516}{0.005}
& \SplitReAlignMeanSD{63.06}{4.34}
& \SplitReAlignMeanSD{0.00}{0.00}
& \SplitReAlignMeanSD{61.04}{3.99}
& 11.19
& \SplitReAlignMeanSD{0.512}{0.038}
& \SplitReAlignMeanSD{43.78}{5.83}
& \SplitReAlignMeanSD{0.00}{0.00}
& \SplitReAlignMeanSD{43.27}{6.81}
& 23.54
& \SplitReAlignMeanSD{0.515}{0.030} \\

& MoDe
& \SplitReAlignMeanSD{55.11}{3.15}
& \SplitReAlignMeanSD{0.17}{0.15}
& \SplitReAlignMeanSD{54.36}{2.60}
& 16.12
& \SplitReAlignMeanSD{0.503}{0.022}
& \SplitReAlignMeanSD{69.96}{4.34}
& \SplitReAlignMeanSD{0.00}{0.00}
& \SplitReAlignMeanSD{68.06}{3.31}
& 6.55
& \SplitReAlignMeanSD{0.523}{0.007}
& \SplitReAlignMeanSD{41.01}{3.12}
& \SplitReAlignMeanSD{0.00}{0.00}
& \SplitReAlignMeanSD{40.78}{2.03}
& 25.29
& \SplitReAlignMeanSD{0.526}{0.007} \\

& FedOSD
& \SplitReAlignMeanSD{80.21}{3.85}
& \SplitReAlignMeanSD{79.00}{4.69}
& \SplitReAlignMeanSD{77.16}{2.55}
& 26.43
& \SplitReAlignMeanSD{0.487}{0.040}
& \SplitReAlignMeanSD{80.76}{3.33}
& \SplitReAlignMeanSD{55.07}{23.68}
& \SplitReAlignMeanSD{77.46}{2.32}
& 18.54
& \SplitReAlignMeanSD{0.508}{0.012}
& \SplitReAlignMeanSD{80.91}{3.34}
& \SplitReAlignMeanSD{47.90}{23.44}
& \SplitReAlignMeanSD{77.77}{2.50}
& 16.30
& \SplitReAlignMeanSD{0.519}{0.034} \\

& NoT
& \SplitReAlignMeanSD{74.24}{2.41}
& \SplitReAlignMeanSD{20.93}{7.13}
& \SplitReAlignMeanSD{72.40}{1.60}
& 10.65
& \SplitReAlignMeanSD{0.517}{0.010}
& \SplitReAlignMeanSD{78.95}{2.73}
& \SplitReAlignMeanSD{4.13}{4.57}
& \SplitReAlignMeanSD{76.27}{1.91}
& 2.19
& \SplitReAlignMeanSD{0.525}{0.005}
& \SplitReAlignMeanSD{70.76}{8.05}
& \SplitReAlignMeanSD{2.13}{2.94}
& \SplitReAlignMeanSD{68.80}{7.42}
& 6.75
& \SplitReAlignMeanSD{0.512}{0.024} \\

& FedCARE
& \SplitReAlignMeanSD{80.53}{3.71}
& \SplitReAlignMeanSD{72.47}{5.97}
& \SplitReAlignMeanSD{77.41}{2.55}
& 24.27
& \SplitReAlignMeanSD{0.508}{0.029}
& \SplitReAlignMeanSD{81.26}{2.93}
& \SplitReAlignMeanSD{35.43}{13.50}
& \SplitReAlignMeanSD{78.37}{2.27}
& 12.47
& \SplitReAlignMeanSD{0.526}{0.035}
& \SplitReAlignMeanSD{81.29}{2.98}
& \SplitReAlignMeanSD{36.07}{13.20}
& \SplitReAlignMeanSD{78.38}{2.27}
& 12.69
& \SplitReAlignMeanSD{0.525}{0.042} \\

& FedMemPrune
& \SplitReAlignMeanSD{73.20}{5.65}
& \SplitReAlignMeanSD{19.77}{5.08}
& \SplitReAlignMeanSD{71.31}{4.60}
& 10.97
& \SplitReAlignMeanSD{0.511}{0.007}
& \SplitReAlignMeanSD{78.29}{4.70}
& \SplitReAlignMeanSD{0.00}{0.00}
& \SplitReAlignMeanSD{75.48}{3.73}
& 1.30
& \SplitReAlignMeanSD{0.530}{0.011}
& \SplitReAlignMeanSD{72.54}{5.34}
& \SplitReAlignMeanSD{0.00}{0.00}
& \SplitReAlignMeanSD{70.67}{3.74}
& 4.82
& \SplitReAlignMeanSD{0.493}{0.014} \\

\addlinespace[2pt]

& SURE
& \SplitReAlignMeanSD{82.37}{3.01}
& \SplitReAlignMeanSD{15.20}{3.44}
& \SplitReAlignMeanSD{78.93}{2.18}
& 6.28
& \SplitReAlignMeanSD{0.532}{0.025}
& \SplitReAlignMeanSD{82.15}{2.96}
& \SplitReAlignMeanSD{1.47}{2.25}
& \SplitReAlignMeanSD{78.82}{2.01}
& 1.59
& \SplitReAlignMeanSD{0.539}{0.033}
& \SplitReAlignMeanSD{81.38}{2.49}
& \SplitReAlignMeanSD{5.47}{4.72}
& \SplitReAlignMeanSD{78.30}{2.35}
& 2.50
& \SplitReAlignMeanSD{0.527}{0.033} \\

\rowcolor{gray!12}
& \textbf{SplitReAlign}
& \SplitReAlignMeanSD{80.25}{3.04}
& \SplitReAlignMeanSD{4.67}{0.68}
& \SplitReAlignMeanSD{77.39}{2.51}
& \textbf{1.58}
& \SplitReAlignMeanSD{0.508}{0.004}
& \SplitReAlignMeanSD{81.05}{3.03}
& \SplitReAlignMeanSD{0.10}{0.10}
& \SplitReAlignMeanSD{78.25}{2.11}
& \textbf{0.58}
& \SplitReAlignMeanSD{0.510}{0.014}
& \SplitReAlignMeanSD{81.83}{3.18}
& \SplitReAlignMeanSD{1.67}{1.50}
& \SplitReAlignMeanSD{78.59}{2.18}
& \textbf{1.47}
& \SplitReAlignMeanSD{0.527}{0.015} \\

\midrule[1pt]

% ===================== CIFAR-100 =====================

\multirow{10}{*}{\rotatebox[origin=c]{90}{\shortstack[c]{CIFAR-100\\Dirichlet ($\alpha=0.1$)}}}
& Retrain
& \SplitReAlignMeanSD{64.02}{2.98}
& \SplitReAlignMeanSD{0.00}{0.00}
& \SplitReAlignMeanSD{51.98}{0.83}
& 0.00
& \SplitReAlignMeanSD{0.516}{0.032}
& \SplitReAlignMeanSD{64.02}{2.98}
& \SplitReAlignMeanSD{0.00}{0.00}
& \SplitReAlignMeanSD{51.98}{0.83}
& 0.00
& \SplitReAlignMeanSD{0.516}{0.032}
& \SplitReAlignMeanSD{64.02}{2.98}
& \SplitReAlignMeanSD{0.00}{0.00}
& \SplitReAlignMeanSD{51.98}{0.83}
& 0.00
& \SplitReAlignMeanSD{0.516}{0.032} \\

\cmidrule(l){2-17}

& Fine-tune
& \SplitReAlignMeanSD{66.57}{3.07}
& \SplitReAlignMeanSD{59.33}{7.09}
& \SplitReAlignMeanSD{52.58}{1.14}
& 20.83
& \SplitReAlignMeanSD{0.505}{0.023}
& \SplitReAlignMeanSD{68.17}{3.20}
& \SplitReAlignMeanSD{17.33}{11.02}
& \SplitReAlignMeanSD{52.58}{1.10}
& 7.36
& \SplitReAlignMeanSD{0.464}{0.011}
& \SplitReAlignMeanSD{68.32}{3.19}
& \SplitReAlignMeanSD{16.00}{11.27}
& \SplitReAlignMeanSD{52.62}{1.04}
& 6.98
& \SplitReAlignMeanSD{0.496}{0.030} \\

& SalUn
& \SplitReAlignMeanSD{61.14}{4.62}
& \SplitReAlignMeanSD{13.67}{6.81}
& \SplitReAlignMeanSD{49.05}{2.35}
& 6.49
& \SplitReAlignMeanSD{0.528}{0.074}
& \SplitReAlignMeanSD{46.09}{3.16}
& \SplitReAlignMeanSD{0.00}{0.00}
& \SplitReAlignMeanSD{37.95}{1.73}
& 10.65
& \SplitReAlignMeanSD{0.535}{0.039}
& \SplitReAlignMeanSD{46.20}{3.65}
& \SplitReAlignMeanSD{0.00}{0.00}
& \SplitReAlignMeanSD{39.41}{2.07}
& 10.13
& \SplitReAlignMeanSD{0.503}{0.047} \\

& MoDe
& \SplitReAlignMeanSD{56.13}{2.45}
& \SplitReAlignMeanSD{0.33}{0.58}
& \SplitReAlignMeanSD{46.09}{2.14}
& 4.70
& \SplitReAlignMeanSD{0.426}{0.044}
& \SplitReAlignMeanSD{57.18}{4.08}
& \SplitReAlignMeanSD{0.00}{0.00}
& \SplitReAlignMeanSD{47.26}{1.72}
& 3.85
& \SplitReAlignMeanSD{0.462}{0.028}
& \SplitReAlignMeanSD{48.78}{4.14}
& \SplitReAlignMeanSD{0.00}{0.00}
& \SplitReAlignMeanSD{42.40}{2.05}
& 8.27
& \SplitReAlignMeanSD{0.471}{0.060} \\

& FedOSD
& \SplitReAlignMeanSD{65.81}{3.01}
& \SplitReAlignMeanSD{56.33}{1.15}
& \SplitReAlignMeanSD{52.09}{1.23}
& 19.41
& \SplitReAlignMeanSD{0.490}{0.048}
& \SplitReAlignMeanSD{65.82}{3.12}
& \SplitReAlignMeanSD{3.00}{1.00}
& \SplitReAlignMeanSD{52.07}{1.17}
& 1.63
& \SplitReAlignMeanSD{0.514}{0.053}
& \SplitReAlignMeanSD{65.74}{3.05}
& \SplitReAlignMeanSD{3.00}{0.00}
& \SplitReAlignMeanSD{52.09}{1.23}
& 1.61
& \SplitReAlignMeanSD{0.549}{0.082} \\

& NoT
& \SplitReAlignMeanSD{26.26}{5.79}
& \SplitReAlignMeanSD{7.33}{4.73}
& \SplitReAlignMeanSD{23.72}{4.76}
& 24.45
& \SplitReAlignMeanSD{0.544}{0.095}
& \SplitReAlignMeanSD{64.00}{3.69}
& \SplitReAlignMeanSD{17.00}{7.94}
& \SplitReAlignMeanSD{50.65}{1.09}
& 6.12
& \SplitReAlignMeanSD{0.509}{0.066}
& \SplitReAlignMeanSD{53.41}{2.84}
& \SplitReAlignMeanSD{1.00}{1.73}
& \SplitReAlignMeanSD{45.26}{0.67}
& 6.11
& \SplitReAlignMeanSD{0.524}{0.048} \\

& FedCARE
& \SplitReAlignMeanSD{65.87}{2.83}
& \SplitReAlignMeanSD{67.33}{5.13}
& \SplitReAlignMeanSD{52.30}{1.08}
& 23.17
& \SplitReAlignMeanSD{0.529}{0.050}
& \SplitReAlignMeanSD{66.46}{2.72}
& \SplitReAlignMeanSD{22.67}{12.90}
& \SplitReAlignMeanSD{52.55}{1.13}
& 8.56
& \SplitReAlignMeanSD{0.524}{0.064}
& \SplitReAlignMeanSD{66.43}{2.73}
& \SplitReAlignMeanSD{22.67}{12.90}
& \SplitReAlignMeanSD{52.53}{1.16}
& 8.54
& \SplitReAlignMeanSD{0.528}{0.055} \\

& FedMemPrune
& \SplitReAlignMeanSD{20.04}{14.00}
& \SplitReAlignMeanSD{19.00}{7.81}
& \SplitReAlignMeanSD{18.27}{12.24}
& 32.23
& \SplitReAlignMeanSD{0.421}{0.063}
& \SplitReAlignMeanSD{62.45}{3.39}
& \SplitReAlignMeanSD{0.00}{0.00}
& \SplitReAlignMeanSD{49.81}{1.11}
& 1.25
& \SplitReAlignMeanSD{0.536}{0.033}
& \SplitReAlignMeanSD{54.73}{3.29}
& \SplitReAlignMeanSD{0.00}{0.00}
& \SplitReAlignMeanSD{45.84}{1.41}
& 5.15
& \SplitReAlignMeanSD{0.493}{0.036} \\

\addlinespace[2pt]

& SURE
& \SplitReAlignMeanSD{66.40}{2.90}
& \SplitReAlignMeanSD{29.67}{6.66}
& \SplitReAlignMeanSD{52.67}{1.02}
& 10.91
& \SplitReAlignMeanSD{0.493}{0.040}
& \SplitReAlignMeanSD{67.70}{3.04}
& \SplitReAlignMeanSD{10.00}{9.12}
& \SplitReAlignMeanSD{52.43}{0.99}
& 4.71
& \SplitReAlignMeanSD{0.459}{0.006}
& \SplitReAlignMeanSD{68.30}{3.21}
& \SplitReAlignMeanSD{18.00}{12.77}
& \SplitReAlignMeanSD{52.69}{1.08}
& 7.66
& \SplitReAlignMeanSD{0.475}{0.037} \\

\rowcolor{gray!12}
& \textbf{SplitReAlign}
& \SplitReAlignMeanSD{64.74}{3.48}
& \SplitReAlignMeanSD{2.00}{1.73}
& \SplitReAlignMeanSD{51.46}{1.15}
& \textbf{1.08}
& \SplitReAlignMeanSD{0.479}{0.115}
& \SplitReAlignMeanSD{64.88}{3.89}
& \SplitReAlignMeanSD{0.00}{0.00}
& \SplitReAlignMeanSD{51.29}{1.55}
& \textbf{0.52}
& \SplitReAlignMeanSD{0.511}{0.055}
& \SplitReAlignMeanSD{66.18}{4.55}
& \SplitReAlignMeanSD{0.00}{0.00}
& \SplitReAlignMeanSD{51.60}{1.63}
& \textbf{0.85}
& \SplitReAlignMeanSD{0.481}{0.050} \\

\bottomrule
\end{tabular}%
}

\vspace{2pt}
\begin{minipage}{\textwidth}
\footnotesize
\emph{Note.} Bold indicates the best AG.
SplitReAlign uses C-guided in Server-only and Standard otherwise.
\end{minipage}
\end{table*}

\paragraph{Training and baselines.}
The Original and Retrain models are trained for 200 global rounds, with one local epoch per round. All unlearning methods start from the same Original checkpoint and run for 20 rounds. We compare against SURE~\cite{chen2026sure}, developed in the first systematic study of SFU, alongside conventional unlearning methods (Fine-tune and SalUn~\cite{fan2024salun}) and federated unlearning methods (MoDe~\cite{zhao2024mode}, FedOSD~\cite{pan2025fedosd}, NoT~\cite{khalil2025not}, FedCARE~\cite{li2026fedcare}, and FedMemPrune~\cite{wei2026fedmemprune}). Except for SURE, these baselines are adapted to SFL by restricting updates and aggregation to the permitted partitions. Retrain serves as a common reference, training both partitions from scratch on retained data.

% \paragraph{Checkpoint selection and metrics.}
% We select the unlearning checkpoint with the highest retained-class validation accuracy among rounds with class-0 validation accuracy at most $5\%$. If none qualifies, we select the round with the lowest class-0 validation accuracy and flag the run. We report \emph{Retain} (retained-training probe accuracy), \emph{Forget}(class-0 test accuracy), \emph{Test} (macro-average test accuracy over non-target classes), and \emph{MIA AUC} (membership-inference ROC-AUC). Results are reported as mean $\pm$ sample standard deviation over seeds 41, 42, and 43. With $\mathcal{Q}=\{\mathrm{Retain},\mathrm{Forget},\mathrm{Test}\}$,the average gap from Retrain is
% \begin{equation}
% \label{eq:ag}
% \mathrm{AG}
% =\frac{1}{3}\sum_{q\in\mathcal{Q}}
% \left|\bar q_{\mathrm{method}}-\bar q_{\mathrm{Retrain}}\right|,
% \end{equation}
% where $\bar q$ is the seed mean for the same dataset and partition. AG is measured in percentage points; lower is better.

\paragraph{Evaluation protocol.}

We report retained-training accuracy (Retain), target-class test accuracy (Forget), macro-averaged non-target test accuracy (Test), and membership-inference ROC-AUC (MIA). Lower Forget and MIA closer to $0.5$ are preferred, while higher Retain and Test indicate better utility. We select checkpoints maximizing retained-class validation accuracy subject to target-class validation accuracy $\leq 5\%$; otherwise, we choose the checkpoint with the lowest target-class validation accuracy. Results are reported as mean $\pm$ standard deviation over three seeds. We define the average gap (AG) from Retrain as
\begin{equation}
\label{eq:ag}
\resizebox{0.9\columnwidth}{!}{$\displaystyle
\mathrm{AG}=
\frac{1}{3}\sum_{q\in\mathcal{Q}}
\left|\bar q_{\mathrm{method}}-\bar q_{\mathrm{Retrain}}\right|,
\quad
\mathcal{Q}=\{\mathrm{Retain},\mathrm{Forget},\mathrm{Test}\},
$}
\end{equation}
where $\bar q$ is the seed-averaged accuracy for metric $q$. Lower AG indicates greater agreement with Retrain.

\subsection{Main Results}
\label{sec:main_classwise}

% \paragraph{Overall comparison.}
% Table~\ref{tab:class_wise_main} shows that SplitReAlign
% achieves the lowest AG among all compared methods across
% six settings, indicating close agreement with Retrain.
% Server-only consistently achieves the lowest AG among
% our three update modes on both datasets.

% Across update modes, Client-only effectively suppresses
% the forgotten class while preserving retained-class utility,
% whereas Server-only achieves a more favorable
% forgetting--utility trade-off even with the client partition
% frozen. Joint updates also outperform the baselines but
% do not improve upon Server-only, suggesting that greater
% update flexibility does not necessarily yield better
% unlearning performance.

% \subsection{Main Results: Class-wise Unlearning}
% \label{sec:main_classwise}

\paragraph{Overall comparison.}
Table~\ref{tab:class_wise_main} shows that SplitReAlign achieves the lowest AG across all six dataset--update-mode configurations. Compared with SURE, SplitReAlign reduces AG from 6.28 to 1.58 in Client-only on CIFAR-10 and from 10.91 to 1.08 on CIFAR-100, with consistent improvements in Server-only and Joint. Importantly, these gains reflect a better forgetting--utility balance rather than forgetting alone. For example, on CIFAR-10 under Client-only updates, Fine-tune retains substantial target-class accuracy (44.47\%), whereas SalUn suppresses the target class (0.97\%) at the cost of retained-class test accuracy (54.96\%). SplitReAlign instead achieves 4.67\% forgetting accuracy while maintaining 77.39\% test accuracy, close to Retrain (77.44\%).

\paragraph{Impact of partition constraints.}
Server-only achieves the lowest AG among the three update modes on both datasets (0.58 on CIFAR-10 and 0.52 on CIFAR-100), despite keeping the client partition frozen. Joint updates also consistently outperform SURE but do not surpass Server-only in AG. These findings suggest that effective unlearning depends not simply on the number of trainable parameters, but on how representation guidance is adapted to the permitted update paths. SplitReAlign also yields MIA AUC values near 0.5 across the evaluated configurations.

% ============================================================
% 5.3 Representation Visualization
% ============================================================

% \subsection{Representation Visualization}
% \label{sec:representation_visualization}

% Figure~\ref{fig:server-representation} shows server-side
% representations of the same validation images before
% and after Client-only unlearning, projected onto a
% fixed PCA basis fitted on Original-model features.
% After unlearning, forgotten-class features exhibit
% greater overlap with the selected anchor region
% in the projected space, while retained-class
% features largely preserve their distribution.
% Although server parameters remain frozen,
% client updates alter downstream representations
% through changes in cut-layer activations.

\subsection{Representation Realignment Analysis}
\label{sec:representation_visualization}

\begin{figure}[t]
    \centering
    \includegraphics[width=\columnwidth]{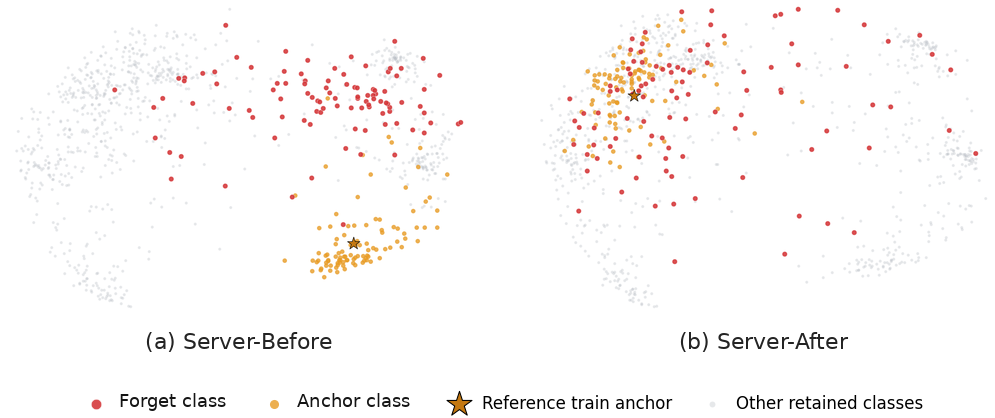}
    % \caption{\textbf{Server-side representations before and after unlearning on CIFAR-10 under Dirichlet ($\alpha=0.1$), projected onto a fixed 2D PCA basis fitted on Original-model features. Red and orange points denote the forgotten and selected anchor classes, the star marks the reference training anchor, and gray points denote other retained classes.}}
    \caption{Server-side representation realignment under Client-only unlearning on CIFAR-10. The same validation samples are projected onto a fixed PCA basis fitted on the original model. Red, orange, and gray denote the forgotten class, selected anchor class, and other retained classes, respectively. The star marks the reference training anchor.}
    \label{fig:server-representation}

\end{figure}

We visualize server-side representations before and after Client-only unlearning using fixed PCA (Fig.~\ref{fig:server-representation}). Forgotten-class features shift toward the retained-class anchor, increasing overlap while largely preserving other retained classes. Despite frozen server parameters, client-side updates reshape server representations through the cut layer. This highlights how cross-partition guidance enables selective realignment with limited disruption to retained features.

% ============================================================
% 5.4 Ablation Study
% ============================================================

% \subsection{Ablation Study}
% \label{sec:ablation}

% Table~\ref{tab:endpoint_representation_ablation}
% examines endpoint guidance under different update
% constraints. In Client-only, dual-endpoint guidance
% achieves lower AG than either single-endpoint loss on
% both datasets. In Server-only, client-guided weighting
% improves upon server-side guidance alone. In Joint,
% adding client-side guidance marginally improves AG
% on CIFAR-100 but worsens it on CIFAR-10, suggesting
% that complementary endpoint guidance is most beneficial
% under restricted parameter updates.

\subsection{Ablation Study}
\label{sec:ablation}

To assess partition-aware guidance, we compare SplitReAlign with single-endpoint variants (Table~\ref{tab:endpoint_representation_ablation}). In Client-only, dual-endpoint guidance achieves lower AG, highlighting the benefit of supervision through the frozen server. In Server-only, client-guided weighting reduces AG from 0.70 to 0.58 on CIFAR-10 and from 0.58 to 0.52 on CIFAR-100. Joint updates yield mixed results, improving CIFAR-100 but degrading CIFAR-10 relative to server-side guidance alone. These findings highlight the value of endpoint coordination under restricted updates.

\section{Conclusion}
\label{sec:conclusion}

SplitReAlign demonstrates that effective unlearning in split federated systems need not depend on unrestricted parameter updates. By exploiting representation dependencies across model partitions, even frozen components can provide meaningful guidance for selective forgetting. More broadly, our findings shift the focus of partition-constrained unlearning from parameter accessibility to representation coordination, offering a promising direction for responsible knowledge removal in distributed AI systems.

\begin{table}[t]
\centering
\caption{Ablation of partition-aware representation guidance.}
\label{tab:endpoint_representation_ablation}

\begingroup
\scriptsize
\setlength{\tabcolsep}{2pt}
\renewcommand{\arraystretch}{1.05}
\setlength{\aboverulesep}{0.6pt}
\setlength{\belowrulesep}{0.6pt}

\resizebox{\columnwidth}{!}{%
\begin{tabular}{@{}l*{12}{r}@{}}
\toprule
& \multicolumn{6}{c}{CIFAR-10}
& \multicolumn{6}{c}{CIFAR-100} \\
\cmidrule(lr){2-7}
\cmidrule(lr){8-13}

& \multicolumn{2}{c}{C-only}
& \multicolumn{2}{c}{S-only}
& \multicolumn{2}{c}{Joint}
& \multicolumn{2}{c}{C-only}
& \multicolumn{2}{c}{S-only}
& \multicolumn{2}{c}{Joint} \\

Variant
& AG$\downarrow$ & MIA
& AG$\downarrow$ & MIA
& AG$\downarrow$ & MIA
& AG$\downarrow$ & MIA
& AG$\downarrow$ & MIA
& AG$\downarrow$ & MIA \\
\midrule

\rowcolor{gray!12}
\textbf{Ours}
& \textbf{1.58} & 0.508
& \textbf{0.58} & 0.510
& 1.47 & 0.527
& \textbf{1.08} & 0.479
& \textbf{0.52} & 0.511
& \textbf{0.85} & 0.481 \\

Only-$\mathcal{L}_C$
& 16.13 & 0.526
& -- & --
& 3.96 & 0.539
& 16.74 & 0.529
& -- & --
& 5.07 & 0.466 \\

Only-$\mathcal{L}_S$
& 4.45 & 0.501
& 0.70 & 0.536
& \textbf{0.92} & 0.529
& 1.18 & 0.518
& 0.58 & 0.529
& 0.89 & 0.543 \\

\bottomrule
\end{tabular}%
}
\endgroup

\vspace{1pt}
\parbox{\columnwidth}{%
\scriptsize
\emph{Note.} C-only and S-only denote Client-only and Server-only, respectively.
Bold indicates the lowest AG within each setting.
Server-only Ours uses client-guided weighting.
}
\end{table}

\bibliographystyle{ACM-Reference-Format}
\bibliography{references}

@misc{chen2026sure,
  author       = {Jia Chen and Yecheng Lv and Xinrong Hu
                  and Mingdi Huang and Jun Bai},
  title        = {Split Federated Unlearning},
  year         = {2026},
  eprint       = {2610.36465},
  eprinttype   = {arXiv},
  eprintclass  = {cs.LG},
  url          = {https://arxiv.org/abs/2610.36465}
}

@inproceedings{fan2024salun,
  author    = {Chongyu Fan and Jiancheng Liu and Yihua Zhang
               and Eric Wong and Dennis Wei and Sijia Liu},
  title     = {{SalUn}: Empowering Machine Unlearning via
               Gradient-based Weight Saliency in Both Image
               Classification and Generation},
  booktitle = {International Conference on Learning Representations (ICLR)},
  year      = {2024},
  url       = {https://proceedings.iclr.cc/paper_files/paper/2024/hash/ec4d2e436794d1bf55ca83f5ebb31887-Abstract-Conference.html}
}

@article{hu2019adaptive,
  author       = {Hanzhang Hu and Debadeepta Dey and Martial Hebert
                  and J. Andrew Bagnell},
  title        = {Learning Anytime Predictions in Neural Networks
                  via Adaptive Loss Balancing},
  journal      = {Proceedings of the AAAI Conference
                  on Artificial Intelligence},
  volume       = {33},
  number       = {1},
  pages        = {3812--3821},
  year         = {2019},
  doi          = {10.1609/aaai.v33i01.33013812}
}

@inproceedings{khalil2025not,
  author       = {Yasser H. Khalil and Leo Brunswic
                  and Soufiane Lamghari and Xu Li
                  and Mahdi Beitollahi and Xi Chen},
  title        = {{NoT}: Federated Unlearning via Weight Negation},
  booktitle    = {Proceedings of the IEEE/CVF Conference on
                  Computer Vision and Pattern Recognition (CVPR)},
  pages        = {25759--25769},
  year         = {2025},
  url          = {https://openaccess.thecvf.com/content/CVPR2025/html/
                  Khalil_NoT_Federated_Unlearning_via_Weight_Negation_CVPR_2025_paper.html}
}

@inproceedings{li2026fedcare,
  author       = {Yue Li and Mingmin Chu and Xilei Yang
                  and Da Xiao and Ziqi Xu and Wei Shao
                  and Qipeng Song and Hui Li},
  title        = {{FedCARE}: Federated Unlearning with
                  Conflict-Aware Projection and
                  Relearning-Resistant Recovery},
  booktitle    = {Proceedings of the Thirty-Fifth International
                  Joint Conference on Artificial Intelligence
                  (IJCAI 2026)},
  pages        = {5576--5584},
  year         = {2026},
  doi          = {10.24963/ijcai.2026/621}
}

@article{pan2025fedosd,
  author       = {Zibin Pan and Zhichao Wang and Chi Li
                  and Kaiyan Zheng and Boqi Wang
                  and Xiaoying Tang and Junhua Zhao},
  title        = {Federated Unlearning with Gradient Descent
                  and Conflict Mitigation},
  journal      = {Proceedings of the AAAI Conference
                  on Artificial Intelligence},
  volume       = {39},
  number       = {19},
  pages        = {19804--19812},
  year         = {2025},
  doi          = {10.1609/aaai.v39i19.34181}
}

@inproceedings{wei2026fedmemprune,
  author       = {Jiaheng Wei and Yanjun Zhang and He Zhang
                  and Leo Yu Zhang and Chao Chen
                  and Kok-Leong Ong and Jun Zhang
                  and Yang Xiang},
  title        = {Rethinking Federated Unlearning via
                  the Lens of Memorization},
  booktitle    = {Proceedings of the 32nd ACM SIGKDD Conference
                  on Knowledge Discovery and Data Mining V.2
                  (KDD 2026)},
  address      = {Jeju Island, Republic of Korea},
  publisher    = {Association for Computing Machinery},
  numpages     = {23},
  year         = {2026},
  doi          = {10.1145/3770855.3817785}
}

@article{zhao2024mode,
  author       = {Yian Zhao and Pengfei Wang and Heng Qi
                  and Jianguo Huang and Zongzheng Wei
                  and Qiang Zhang},
  title        = {Federated Unlearning With Momentum Degradation},
  journal      = {IEEE Internet of Things Journal},
  volume       = {11},
  number       = {5},
  pages        = {8860--8870},
  year         = {2024},
  doi          = {10.1109/JIOT.2023.3321594}
}

@inproceedings{thapa2022splitfed,
  title={Splitfed: When federated learning meets split learning},
  author={Thapa, Chandra and Arachchige, Pathum Chamikara Mahawaga and Camtepe, Seyit and Sun, Lichao},
  booktitle={Proceedings of the AAAI conference on artificial intelligence},
  volume={36},
  number={8},
  pages={8485--8493},
  year={2022}
}

@InProceedings{gao2026illusion,
  title = 	 { An Illusion of Unlearning? Assessing Machine Unlearning Through Internal Representations },
  author =       {Gao, Yichen and Unal, Altay and Rangamani, Akshay and Zhu, Zhihui},
  booktitle = 	 {Proceedings of The 29th International Conference on Artificial Intelligence and Statistics},
  pages = 	 {4114--4122},
  year = 	 {2026},
  editor = 	 {Khan, Emtiyaz and Li, Yingzhen and Solin, Arno and Ramdas, Aaditya},
  volume = 	 {300},
  series = 	 {Proceedings of Machine Learning Research},
  month = 	 {02--05 May},
  publisher =    {PMLR},
  url = 	 {https://proceedings.mlr.press/v300/gao26e.html},
}

@inproceedings{li2026cross,
  title={Cross-modal unlearning via influential neuron path editing in multimodal large language models},
  author={Li, Kunhao and Li, Wenhao and Wu, Di and Yang, Lei and Bai, Jun and Jia, Ju and Xue, Jason},
  booktitle={Proceedings of the AAAI Conference on Artificial Intelligence},
  volume={40},
  number={42},
  pages={35589--35597},
  year={2026}
}

\end{document}